\documentclass{article}
\usepackage{spconf,amsmath,graphicx}

\usepackage{enumitem}
\setlist{nosep, leftmargin=14pt}

\usepackage{mwe} 
\usepackage{multirow}
\usepackage{booktabs}
\usepackage{tabularx}
\usepackage{arydshln} 
\usepackage{array}

\usepackage{fancyhdr}
\usepackage{lipsum}
\title{Surgical Scene Segmentation by Transformer \\ With Asymmetric Feature Enhancement}
\name{Cheng Yuan$^{1}$, Yutong Ban$^{1,*}$\thanks{$^{*}$Corresponding email: yban@sjtu.edu.cn}}
\address{$^{1}$UM-SJTU Joint Institute, Shanghai Jiao Tong University, Shanghai, China}
\begin{document}
%
\maketitle
\begin{abstract}
Surgical scene segmentation is a fundamental task for robotic-assisted laparoscopic surgery understanding. It often contains various anatomical structures and surgical instruments, where similar local textures and fine-grained structures make the segmentation a difficult task. Vision-specific transformer method is a promising way for surgical scene understanding. However, there are still two main challenges. Firstly, the absence of inner-patch information fusion leads to poor segmentation performance. Secondly, the specific characteristics of anatomy and instruments are not specifically modeled. To tackle the above challenges, we propose a novel Transformer-based framework with an Asymmetric Feature Enhancement module (TAFE), which enhances local information and then actively fuses the improved feature pyramid into the embeddings from transformer encoders by a multi-scale interaction attention strategy. The proposed method outperforms the SOTA methods in several different surgical segmentation tasks and additionally proves its ability of fine-grained structure recognition. Code is available at https://github.com/cyuan-sjtu/ViT-asym.
\end{abstract}
\section{Introduction}
\label{sec:typestyle}

\begin{figure}[t]
\begin{minipage}[b]{.48\linewidth}
  \centering
  \centerline{\includegraphics[width=4.0cm]{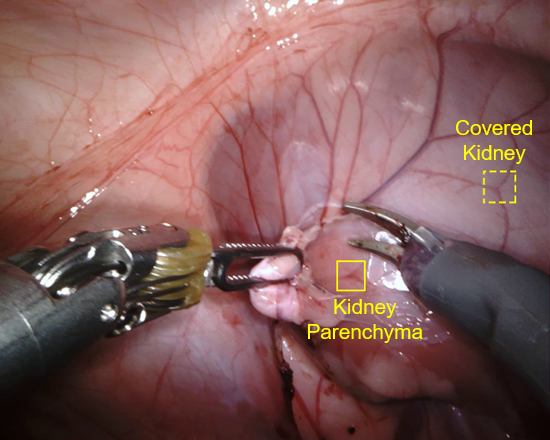}}
  \centerline{(a) Local Feature Similarity}\medskip
\end{minipage}
\hfill
\begin{minipage}[b]{0.48\linewidth}
  \centering
  \centerline{\includegraphics[width=4.0cm]{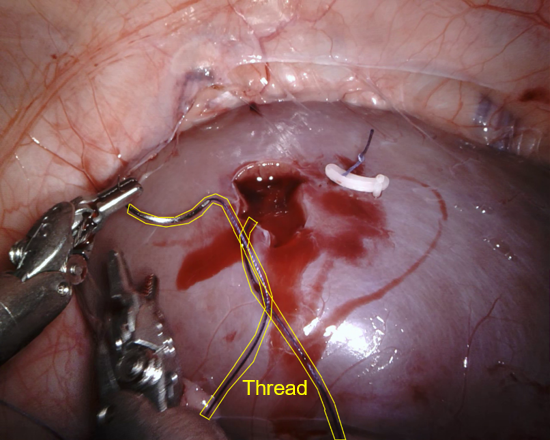}}
  \centerline{(b) Structure Complexity}\medskip
\end{minipage}
\vspace{-2ex}
\caption{Challenges in surgical scene segmentation: (a) local feature similarity between covered kidney and kidney parenchyma; (b) fine-grained structure complexity of tubular instrument, such as thread.}
\label{fig1}
\end{figure}

Minimally invasive laparoscopic surgery has been a common clinical application, showing effective reduction of trauma and recovery period. Achieving accurate surgical scene segmentation can facilitate further pose estimation \cite{allan20183}, surgery navigation \cite{collins2020augmented}, and risk assessment \cite{ali2021deep}. However, accurate segmentation of surgical scene is a challenging task, which is due to two reasons. Firstly, different from non-surgical scene, local features of various anatomies and instruments in surgical scene exhibit high similarity, as shown in Fig.~\ref{fig1}(a). Secondly, complex intersections among multiple adjacent regions make the boundaries blurred and tortuous, especially as the fine-grained structures as shown in Fig.~\ref{fig1}(b).

\begin{figure*}[t]

  \centering
  \centerline{\includegraphics[width=\textwidth]{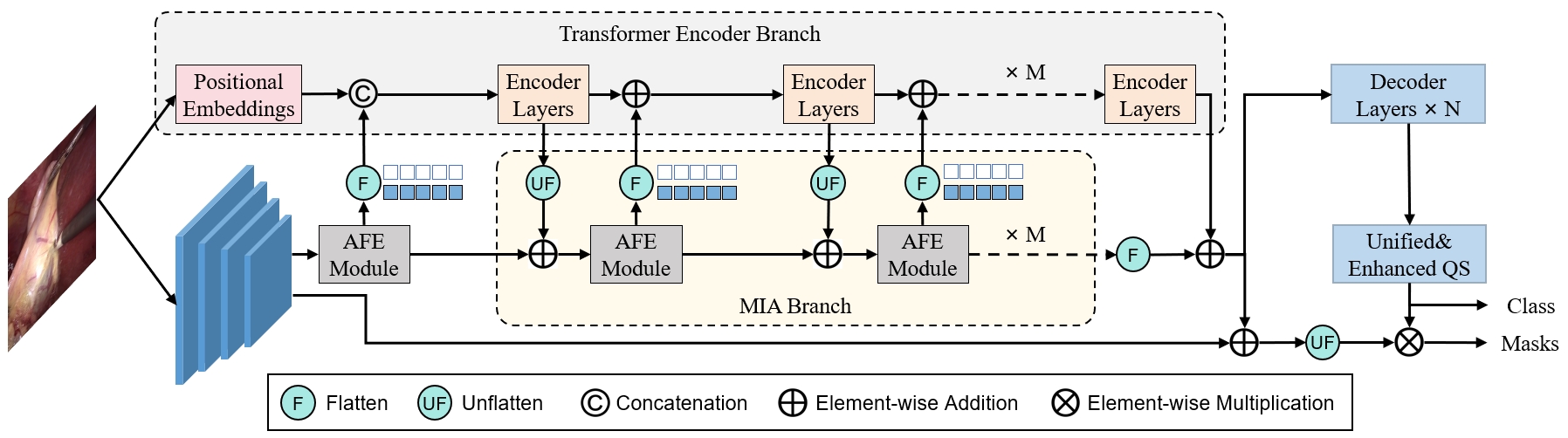}}
%
\caption{The overall architecture of \textbf{TAFE}. It contains a transformer encoder-decoder backbone injected with the \textbf{M}ulti-scale \textbf{I}nteraction \textbf{A}ttention (MIA) branch and the \textbf{A}symmetric \textbf{F}eature \textbf{E}nhancement (AFE) module.}
\label{fig2}
\end{figure*}

Works have done to improve the segmentation of both anatomical structures and surgical instruments. One important category is specific-designed convolution networks, leveraging its local continuity and multiscale capability to the most extend. Yang et al. \cite{yang2020deeplab_v3_plus} used atrous convolution to obtain multiscale context features and achieved good performance in robotic scene segmentation. Ni et al. \cite{ni2022space} designed a space squeeze reasoning network with a low-rank bilinear feature fusion strategy to improve the recognition ability of different surgical regions. Recently, Liu et al. \cite{liu2023lskanet} proposed a long strip kernel attention network and achieved the best results in robotic surgical scene segmentation. The other category is inspired by the success of transformer and applied into dense prediction tasks such as detection and segmentation. As the pioneer, Nicolas et al. \cite{carion2020end} proposed an end-to-end query-based transformer object detector and achieved objective set-prediction by bipartite matching. Jin et al. \cite{jin2022exploring} constructed a transformer-based framework that explored intra- and inter-video relations to capture global context and boost surgical scene segmentation performance. Similarly, Mask2Former \cite{cheng2021per} and MaskDINO \cite{li2023mask} used query-based transformer architectures to perform natural image mask classification. However, these studies commonly ignored the problem of local feature similarity and intersected structure complexity, difficultly recognizing tubular instruments and easily resulting in rough mask and false segmentation.

To overcome the challenges mentioned above, we propose a \textbf{T}ransformer-based framework empowered by specail-designed \textbf{A}symmetric \textbf{F}eature \textbf{E}nhancement, named \textbf{TAFE}, which contains two core parts: the Multiscale Interaction Attention (MIA) branch and the Asymmetric Feature Enhancement (AFE) module. Retaining plain transformer encoder-decoder architecture, this framework can directly load open-source pre-trained weights. MIA actively fuses the enhanced feature pyramid into the embeddings from transformer encoders, which improves both local and global feature representation and reduces the influence of local feature similarity. Inspired by the anisotropic feature perception in medical signal processing \cite{yuan2020ovarian}, AFE is designed to identify specific characteristic of anatomy and instruments. It consists of two blocks separately adopting symmetric convolution for polygon-like feature perception of anatomy and asymmetric convolution for bar-like or tubule-like feature perception of instruments. Our main contributions are as follows:
\begin{itemize}
  \item We propose a multi-scale interaction attention architecture by fusing the enhanced convolution feature pyramid into transformer encoder-decoder backbone . It tackles the challenge of local feature similarity by incorporating multi-scale representations. 
  \item We design an asymmetric feature enhancement module which contains symmetric and asymmetric convolution operations. The symmetric block focus more on polygon-like anatomy, the asymmetric block has better perception on bar-shaped or tubular instruments.
  \item Our proposed method achieves the best overall performance compared with the previous state-of-the-art methods, exhibiting +4.0\% mAP on Endoscapes2023 \cite{murali2023endoscapes} and +11.3\% mIoU on EndoVis2018 \cite{allan20202018}. Moreover, our method outperforms in delicate structure recognition with separate 19.2\% and 4.3\% improvement of no matter hepatocystic triangle dissection or cystic artery.
\end{itemize}

\section{Methodology}
\label{sec:typestyle}

\subsection{Framework Overview}
\label{ssec:subhead}
We propose the surgical scene segmentation framework as shown in Fig.~\ref{fig2}, which includes three parts: (a) a transformer encoder-decoder backbone; (b) the multi-scale interaction attention (MIA) branch; and (c) the asymmetric feature enhancement (AFE) module. Given an image, the multi-scale feature pyramid containing 4 layers with resolutions of 1/4, 1/8, 1/16, and 1/32 is extracted by a stack of convolutions. M-stage feature mutual interactions are constructed to fuse multi-perceptive information from transformer encoders and convolution operations. At each stage, in the transformer encoder branch, the input feature embedding passes through a transformer encoder and is then unflattened to add with the enhanced convolution feature pyramid. Meanwhile, in the MIA branch, the input feature pyramid enhanced by the AFE module is further flattened to add with the output feature embedding from the transformer encoder. After M-stage feature interactions, fused features are fed into the transformer decoder for mask prediction.

\vspace{-2ex}
\subsection{Multi-scale Interaction Attention}
\label{ssec:subhead}
Local feature similarity in surgical scenes usually leads to misidentification, especially when the neural network lacks multi-scale feature reception. To improve the perception capability, we propose a novel multi-perceptive segmentation network architecture with a cross-architecture multi-scale interaction attention branch named MIA. It introduces multi-scale features from convolutions without altering the original transformer backbone, which enables the transformer backbone to utilize the pre-trained model weights. Meanwhile, its mutual information fusion alleviates the lack of inner-patch information interaction in the plain transformer architecture, as well as the problem of global information modeling in the convolution operation.

In order to fuse the transformer encoder feature $F$ and the enhanced multi-scale convolution feature pyramid registered as $\left\{E_1,E_2,E_3,E_4\right\}$. Here, we first unflatten the feature $F$ into 4 corresponding layers by calculating pixel value as:
\begin{equation}
    F_l^{UF}\left(i,j\right)=F\left(i+\left(j-1\right) \ast R_l+\sum_{k=0}^{l-1} R_k^2\right)
\end{equation}
where $l\in\left\{1,2,3,4\right\}$ and $i,j\in\left\{1,2,\dots,R_l\right\}$. $R_l$ represents the size of the $l$-layer feature map in pyramid. Then, each unflattened feature map $F_l^{UF}$ is element-wise added with corresponding enhanced feature map $E_l$ to pass through the next AFE module.

\vspace{-2ex}
\subsection{Asymmetric Feature Enhancement}
\label{ssec:subhead}

From the visual perspective, anatomy and instruments exhibit totally different characteristics, that is irregular polygons for anatomy and regular bars or tubules for instruments. Therefore, we propose a dual-block asymmetric feature enhancement module, named AFE, to focus on different semantic features. The AFE module consists of an anatomy enhancement block and an instrument enhancement block.

\noindent\textbf{Anatomy Enhancement Block:} The input feature map $C_l$ from the convolution pyramid is first passed through a standard convolution with the kernel size of 5$\times$5 as a structural information aggregation step. Next, three symmetric convolution branches are constructed for enhancing network perception of irregular polygon-like anatomy. In order to save computational memory, we adopt a cascaded strip convolution pair to replace the original convolution operation in implementation as follows:
\begin{equation}
    S_{l,m}^{AE}=\mathrm{Conv}_{k_m \times 1}\left(\mathrm{Conv}_{1 \times k_m}\left(C_l^{AE}\right)\right)
\end{equation}
where $C_l^{AE}$ is the aggregated feature map, and the strip kernel $k$ is set to 3, 5, or 7, with $l\in\left\{1,2,3,4\right\}$. 

\noindent\textbf{Instrument Enhancement Block:} Similar to anatomy enhancement, the input feature map $C_l$ is also aggregated at first and passes through three asymmetric convolution branches for improving feature extraction of regular bar-like instruments. We adopt a different topology with a parallel strip convolution pair to learn features in both vertical and horizontal dimensions:
\begin{equation}
    S_{l,m}^{IE}=\mathrm{Conv}_{k_m \times 1}\left(C_l^{IE}\right)+\mathrm{Conv}_{1 \times k_m}\left(C_l^{IE}\right)
\end{equation}
where $C_l^{IE}$ is the same aggregated feature map, and the strip kernel $k$ is also set to 3, 5, or 7, with $l\in\left\{1,2,3,4\right\}$.

After that, the attention map of anatomy or instruments registered as $E_l$ is calculated by summing up from each pyramid layer, adding a shortcut connection, then passing through a 1$\times$1 convolution, and finally multiplying with the aggregated feature map:
\begin{equation}
    E_l=\left(\mathrm{Conv}_{1 \times 1}\left(\sum_{m=0}^2 S_{l,m}^E+C_l^E\right)\right) \otimes C_l^E
\end{equation}
where $S_{l,m}^E$ represents the feature map from the anatomy or instrument enhancement block. Finally, attention maps from two enhancement blocks are summed up as the final output.

\section{Experiments and Results}
\label{sec:typestyle}

\subsection{Datasets}
\label{ssec:subhead}

The experiments were conducted on two laparoscopic surgical scene segmentation datasets and our proposed method was compared with other state-of-the-art (SOTA) methods.

\begin{table*}[t]
    \centering
\caption{Quantitative results on Endoscapes2023. The best results are in bolded, and the second best results are in underlined.}
\label{table1}
  \resizebox{\textwidth}{!}{
    \begin{tabular}{cccccccc}
    \toprule
    \multirow{2}{*}{Method}& \multicolumn{7}{c}{Detection mAP@[0.5:0.95] $\mid$ Segmentation mAP@[0.5:0.95] (\%)}\\
    \cline{2-8}
    {}&  Cystic Plate&  HC Triangle Dissection&  Cystic Artery&  Cystic Duct&  Gallbladder&  Tool&  Overall\\
    \midrule
    Mask-RCNN\cite{he2017mask}& \underline{2.8} $\mid$ 3.3&  2.9 $\mid$ 3.8&  12.7 $\mid$ \underline{11.9}&  7.4 $\mid$ 7.9&  45.8 $\mid$ 59.1& 49.7 $\mid$ 51.2& 20.2 $\mid$ 22.9\\ 
    Cascaded Mask-RCNN\cite{cai2019cascade}&  \textbf{3.1} $\mid$ \textbf{6.5}&  11.1 $\mid$ 6.7&  \underline{15.7} $\mid$ 10.8&  \underline{11.7} $\mid$ 10.0&  62.0 $\mid$ 62.7&  63.2 $\mid$ 56.8& 27.8 $\mid$ 25.6\\
    \hdashline
    Mask2Former\cite{cheng2021per}&  1.4 $\mid$ 1.7&  \underline{14.3} $\mid$ 8.3&  6.4 $\mid$ 7.6&  \textbf{14.7} $\mid$ \textbf{15.9}&  \underline{68.7} $\mid$ 62.6&  \underline{67.1} $\mid$ \textbf{63.5}&  \underline{28.8} $\mid$ \underline{26.6}\\ 
    MaskDINO\cite{li2023mask}&  1.1 $\mid$ 0.5&  12.8 $\mid$ \underline{10.4}&  9.8 $\mid$ 12.2&  7.2 $\mid$ 7.9&  64.0 $\mid$ \textbf{63.2}&  58.7 $\mid$ 59.5&  25.6 $\mid$ 25.6\\ 
    \hdashline
    \textbf{TAFE(Proposed)}&  \underline{2.8} $\mid$ \underline{2.9}&  \textbf{20.1} $\mid$ \textbf{29.6}&  \textbf{20.7} $\mid$ \textbf{16.2}&  11.3 $\mid$ \underline{10.6}& \textbf{70.9} $\mid$ \underline{63.0} &  \textbf{69.5} $\mid$ \underline{61.3}&  \textbf{32.5} $\mid$ \textbf{30.6}\\ 
    \bottomrule
    \end{tabular}}  
\end{table*}

\noindent\textbf{Endoscapes2023:} Endoscapes2023 dataset \cite{murali2023endoscapes} was recently released for surgical scene segmentation, object detection, and critical view of safety assessment. Here, we used its subset consisting of 493 frames from 50 laparoscopic cholecystectomy videos and corresponding detection boxes and segmentation masks, which were divided into 30 training sets, 10 validation sets, and 10 test sets. This segmentation task is defined to recognize required 6 kinds of anatomies and instruments from the whole image.

\noindent\textbf{EndoVis2018:} EndoVis2018 dataset \cite{allan20202018} was released in the 2018 Robot Scene Segmentation Challenge and was made up of 19 sequences which were divided into 15 training sets and 4 test sets. Each frame consists of a stereo pair with 1280$\times$1024 resolution and here we only use the annotated left images. This segmentation task is defined to divide the entire laparoscopic surgical scene into 12 categories including various anatomies and instruments.

\subsection{Implementation Details}
\label{ssec:subhead}
Here, we expound experiment details from three aspects:

\noindent\textbf{Experimental Setup:} The proposed method was implemented in Pytorch (2.2.1 version) and calculated based on two NVIDIA A40 GPUs. Frames of Endoscapes2023 and EndoVis2018 were resized to 512$\times$1024 resolution for memory saving. Model training iteration, training batch size, and base learning rate were set to 20k, 3, and 1e-4, separately.

\noindent\textbf{Evaluation Metric:} We adopted the original evaluation metrics for the respective datasets. For Endoscapes2023 dataset, we used both detection and segmentation mean average precision (mAP@[0.5:0.95]) to keep consistent with the official report. For EndoVis2018 dataset, we used the offical evaluation protocol to calculate mean intersection-over-union (mIoU) and mean Dice coefficient (mDice).

\noindent\textbf{Comparison methods:} We compared the proposed method with several SOTA segmentation methods. On Endoscapes2023 dataset, the compared methods consist of four fine-tuned segmentation models, namely Mask-RCNN \cite{he2017mask}, Cascade Mask-RCNN \cite{cai2019cascade}, Mask2Former \cite{cheng2021per}, and MaskDINO \cite{li2023mask}. On EndoVis2018 dataset, the compared methods include four categories: (1) Three reported methods in the 2018 Robot Scene Segmentation Challenge like NCT, UNC, and OTH \cite{allan20202018}; (2) Multiscale feature fusion networks such as U-Net \cite{ronneberger2015u}, DeepLabv3+ \cite{yang2020deeplab_v3_plus}, UPerNet \cite{xiao2018unified}, and HRNet \cite{wang2020deep}; (3) Fine-tuned Vision transformer models, Mask2Former and MaskDINO; (4) Specific-designed segmentation methods, such as SegFormer \cite{xie2021segformer}, SegNeXt \cite{guo2022segnext}, STswinCL \cite{jin2022exploring}, and LSKANet \cite{liu2023lskanet}.

\subsection{Results and Analysis}
\label{ssec:subhead}

\textbf{Surgical scene segmentation accuracy:} Table \ref{table1} exhibits every category result of Endoscapes2023 dataset and our proposed method achieves the best detection mAP of 32.5\% and segmentation mAP of 30.6\% in overall. Compared with the second best result shown in unederline, difficultly recognized categories such as hepatocystic triangle dissection and cystic artery are separately gets the segmentation mAP improvement of 19.2\% and 4.3\%.

Meanwhile, as shown in Table \ref{table2}, our proposed method achieves the best overall results with the mIoU of 77.5\% and the mDice of 86.6\% on EndoVis2018 dataset. Compared with the previous SOTA method LSKANet, the performance improvement demonstrates the effectiveness of asymmetric feature enhancement. Specially, the mIoU of sequence 2 and 4 outperforms the previous best performance shown in underline with the huge improvement of 19.3\% and 23.5\%. 

\begin{table}[h]
    \centering
\caption{Quantitative results on EndoVis2018. The best results are bolded, and the second best results are underlined.}
\label{table2}
  \resizebox{8.5cm}{!}{
    \begin{tabular}{ccccccc}
    \toprule
    \multirow{2}{*}{Method}& \multirow{2}{*}{mIoU(\%)}& \multicolumn{4}{c}{Squence(mIoU(\%))} &  \multirow{2}{*}{mDice(\%)}\\ 
    \cline{3-6}
    {}&  {}&  Seq1&  Seq2&  Seq3&  Seq4&  {}\\
    \midrule
    NCT\cite{allan20202018}&  58.5&  65.8&  55.5&  76.5&  36.2& /\\ 
    UNC\cite{allan20202018}&  60.7&  66.3&  57.8&  81.4&  37.3& /\\ 
    OTH\cite{allan20202018}&  62.1&  69.1&  57.5&  82.9&  39.0& /\\
    \hdashline
    U-Net\cite{ronneberger2015u}&  50.7&  55.6&  50.5&  69.7&  26.8& 61.5\\ 
    DeepLabv3+\cite{yang2020deeplab_v3_plus}&  58.8&  64.1&  57.0&  82.3&  31.6& 67.3\\ 
    UPerNet\cite{xiao2018unified}& 58.4& 67.4& 54.0& 82.0& 30.2&66.8\\ 
    HRNet\cite{wang2020deep}& 63.3& 68.9& 57.3& 85.0& 42.1&71.8\\
    \hdashline
    Mask2Former\cite{cheng2021per}& 56.1& 58.4& 60.1& 70.5& 35.3& 69.5\\
    MaskDINO\cite{li2023mask}& 58.9& 54.1& \underline{63.8}& 70.8& 47.0& 69.3\\
    \hdashline
    SegFormer\cite{xie2021segformer}& 63.0& 69.1& 55.8& 81.6& 45.5&71.9\\ 
    SegNeXt\cite{guo2022segnext}& 64.3& \textbf{70.6}& 57.1& \underline{84.3}& 45.1&72.5\\ 
    STswinCL\cite{jin2022exploring}& 63.6& 67.0& 63.4& 93.7& 40.3&72.0\\ 
    LSKANet\cite{liu2023lskanet}& \underline{66.2}& 67.5& 63.2& \textbf{85.2}& \underline{48.9}&\underline{75.3}\\
    \hdashline
    \textbf{TAFE(Proposed)}& \textbf{77.5}& \underline{70.5}& \textbf{83.1}& 83.8& \textbf{72.4}&\textbf{86.6}\\
    \bottomrule
    \end{tabular}}
\end{table}

\begin{figure}[h]
  \centering
  \centerline{\includegraphics[width=8.5cm]{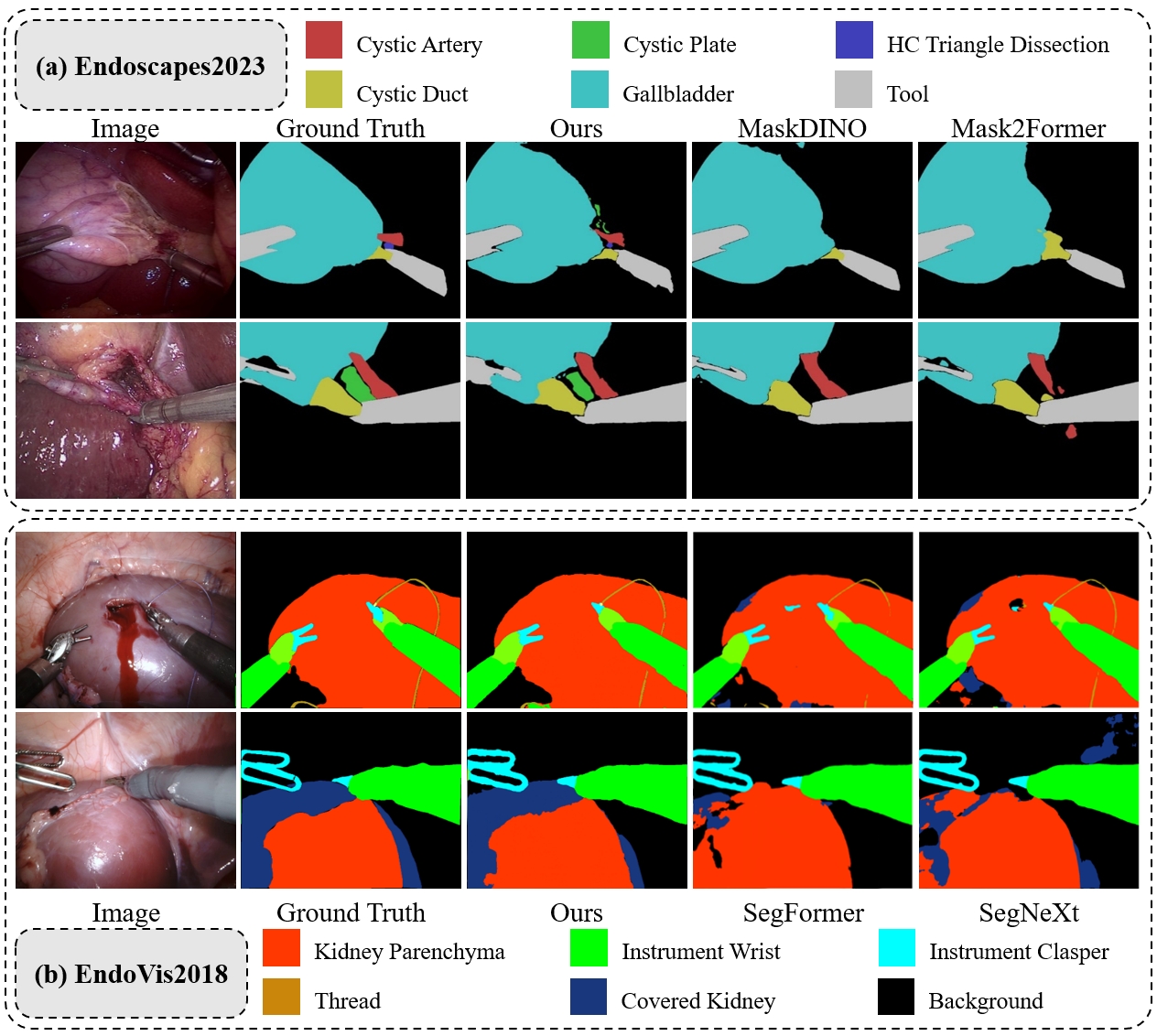}}
%
\vspace{-1ex}
\caption{Visual comparison of the fine-grained structure recognition ability of different methods.}
\label{fig3}
\end{figure}

\noindent\textbf{Fine-grained structure recognition ability:} We explore the ability of the proposed method for fine-grained structure recognition, especially in tubular shape, and carry out a visual comparison of both Endoscapes2023 and EndoVis2018 in Fig. \ref{fig3}. As in Fig. \ref{fig3}(a), dim ambient light causes terrible recognition of all required categories. However, our proposed method still identifies delicate structures such as cystic artery and cystic plate, which are terribly missed in fine-tuned foundation models with good segmentation performance in natural vision field. Besides, as in Fig. \ref{fig3}(b), the slender thread in sample 1 retains a consecutive and complete segmentation result, which is misidentified in other compared methods. In addition, kidney parenchyma and cover kidney have a blurred visual boundary in sample 2, which is more clearly and accurately segmented by our proposed method. In summary, our method shows noticeable advancement in fine-grained structure recognition ability compared with other methods.

\vspace{-4ex}
\section{Conclusion}
\label{sec:typestyle}

In this paper, we proposed a novel surgical scene segmentation transformer framework empowered by a special-designed multi-scale interaction attention architecture, as well as an asymmetric feature enhancement to deal with different feature representations in surgery scene. As a result, the proposed method has achieved the best performance in both scene segmentation on EndoVis2018 and object recognition on Endoscapes2023, which promisingly improve surgical scene comprehensive understanding. In the future work, we plan to further investigate more downstream tasks enabled by the obtained segmentation results. 

\section{Compliance with ethical standards}
\label{sec:ethics}

This research study was conducted retrospectively using human subject data made available in open access by CAMMA Endoscapes. Ethical approval was not required as confirmed by the license attached with the open access data.

\section{Acknowledgments}
\label{sec:acknowledgments}

This work was supported by Shanghai Pujiang Program (Grant No.23PJ1404400).

\bibliographystyle{IEEEbib}
\bibliography{strings,refs}


\vfill
\pagebreak

\end{document}